\documentclass[10pt]{article} %

\usepackage[preprint]{rlj} %

\usepackage{amssymb}            %
\usepackage{mathtools}          %
\usepackage{mathrsfs}           %
\usepackage{graphicx}           %
\usepackage{subcaption}         %
\usepackage[space]{grffile}     %
\usepackage{url}                %
\usepackage{lipsum}             %

\usepackage{algorithm}
\usepackage{algpseudocode}
\usepackage{xcolor}
\usepackage{amsmath,amssymb}
\usepackage{booktabs}
\usepackage{comment}
\usepackage{wrapfig}
\usepackage{multirow}

\definecolor{amber}{HTML}{FFBE28}
\definecolor{flame}{HTML}{E65100}
\definecolor{indigo}{HTML}{535ABF}
\definecolor{cyan}{HTML}{40C1D1}

\definecolor{teal}{HTML}{40888B}
\definecolor{fern}{HTML}{81C784}
\definecolor{mint}{HTML}{E8F5E9}
\definecolor{slate}{HTML}{263238}

\newcommand{\amber}[1]{\noindent{\color{amber}{#1}}}
\newcommand{\flame}[1]{\noindent{\color{flame}{#1}}}
\newcommand{\indigo}[1]{\noindent{\color{indigo}{#1}}}
\newcommand{\cyan}[1]{\noindent{\color{cyan}{#1}}}

\newcommand{\teal}[1]{\noindent{\color{teal}{#1}}}
\newcommand{\fern}[1]{\noindent{\color{fern}{#1}}}

\newcommand{\supp}[1]{Supp.~\ref{#1}}
\renewcommand{\paragraph}[1]{\vspace{.5em}\noindent\textbf{#1.}}

\newcommand{\ba}{\mathbf{a}}

\newcommand{\bs}{\mathbf{s}}

\newcommand{\bu}{\mathbf{u}}

\newcommand{\bz}{\mathbf{z}}

\newcommand{\cJ}{\mathcal{J}}

\DeclareMathOperator*{\argmax}{argmax~}

\makeatletter
\DeclareRobustCommand\onedot{\futurelet\@let@token\@onedot}
\def\@onedot{\ifx\@let@token.\else.\null\fi\xspace}

\makeatother

\hypersetup{
  colorlinks=true,
  urlcolor=flame, 
  linkcolor=flame,
  citecolor=indigo,
  pdfborder={0 0 0},
}

\title{\texttt{Valdi}: Value Diffusion World Models}

\setrunningtitle{\texttt{Valdi}: Value Diffusion World Models}

\author{Christopher Lindenberg\textsuperscript{1}, Kashyap Chitta\textsuperscript{2}}

\emails{{christopher-sebastian.lindenberg@student.uni-tuebingen.de, kashyap@kesai.eu}}

\affiliations{
$^{1}$\textbf{University of Tübingen, Germany}\\
$^{2}$\textbf{KE:SAI -- Kyutai ELLIS Scalable Autonomous Intelligence, Germany}\\
}

\contribution{
    We propose Value Diffusion World Models (\texttt{Valdi}), a TD-MPC-style algorithm, trained end-to-end and online, that uses a latent diffusion model as its dynamics model in place of TD-MPC's MLP.
    }
    {
    Diffusion models are a state-of-the-art approach to world modeling, but are typically applied in pixel space together with model-based RL~\citep{Alonso2024NEURIPS, Hafner2025ARXIV}, where efficiency requirements are not as high as MPC. Online latent planners in the TD-MPC family instead rely on a deterministic MLP dynamics model~\citep{Hansen2022ARXIV, Hansen2023ARXIV}. A latent diffusion dynamics model trained end-to-end inside such a value-based online planning loop has, to our knowledge, not been demonstrated.
    }

\contribution{
    We show that a single diffusion step at both training and inference time is sufficient for \texttt{Valdi} to match the MLP baseline in control performance.
    }
    {
    Diffusion models ordinarily require many iterative denoising steps, the primary reason their inference is considered incompatible with low-latency planning. Whether a single denoising step can retain control performance competitive with a deterministic one-step dynamics model has not, to our knowledge, been examined in the setting of latent MPC.
    }

\contribution{
    Increasing inference-time diffusion steps results in more visually diverse, multimodal predictions but degrades planning, exposing a tension between predictive expressivity and control quality in this setting.
    }
    {
    Diffusion world models are commonly motivated by their capacity to represent multimodal, uncertain futures~\citep{Agarwal2025ArXiv}, with the implicit assumption that richer predictive modeling benefits downstream control. To our knowledge, the effect of increasing this generative expressivity on planning performance within a value-based MPC loop has not been characterized.
    }


\summary{World models can enable inference-time planning by imagining future trajectories, but their practical application requires dynamics prediction that is both fast enough for online use and expressive enough to represent uncertain futures. 
Diffusion models offer a natural mechanism for modeling such uncertain dynamics, yet their iterative inference procedure makes them difficult to combine with low-latency latent planning. 
We introduce Value Diffusion World Models (\texttt{Valdi}), an algorithm that jointly trains a latent diffusion dynamics model with objectives for representation learning, reward prediction, and value prediction in an online control loop.
In preliminary experiments on the CarRacing environment, we find that using a single diffusion step at both training and inference time is sufficient for \texttt{Valdi} to match the control performance of a deterministic MLP dynamics baseline similar to TD-MPC.
However, we show that increasing the inference-time diffusion steps results in more visually diverse, multimodal trajectory predictions, but in turn degrades task performance in this simple environment.
Analysis of the predicted value function suggests that, although diffusion rollouts incur larger short-horizon errors, their value estimates are better grounded than those of the MLP baseline at the longer-horizon rollout depth used for planning.
}

\begin{document}

\makeCover  %
\maketitle  %

\begin{abstract}
World models can enable Model Predictive Control (MPC), but this requires dynamics prediction that is both fast enough for online use and expressive enough to represent uncertain futures. 
Diffusion models offer a natural mechanism for modeling uncertain dynamics, yet their iterative inference procedure makes them difficult to use for low-latency latent planning. 
We bridge this gap with Value Diffusion World Models (\texttt{Valdi}), combining end-to-end online training for MPC with a latent diffusion dynamics model. 
In preliminary experiments on the CarRacing environment, we show that \texttt{Valdi}, using a single diffusion step at both training and inference, matches a deterministic MLP baseline.
Our experiments expose a trade-off between predictive multimodality and control performance in this setup.
Code is available at \href{https://github.com/Kit115/ValueDiffusionWorldModels}{https://github.com/Kit115/ValueDiffusionWorldModels}.
\end{abstract}

\section{Introduction}
\label{sec:intro}

Learned models of environment dynamics, also called \textit{world models}, enable inference-time planning methods such as Model Predictive Control (MPC)~\citep{Garcia1989}. To run online, MPC requires rapid inference, favoring lightweight \textit{latent} world models that predict compressed representations of observations rather than raw pixels~\citep{Hansen2022ARXIV}. Latent world models, however, are prone to representation collapse due to trivial solutions in their optimization space~\citep{Maes2026ARXIV}.

\begin{wrapfigure}[16]{r}{0.50\textwidth}
\centering
\vspace{-0.2em}
\begin{subfigure}[t]{0.49\linewidth}
    \centering
    \includegraphics[width=\linewidth]{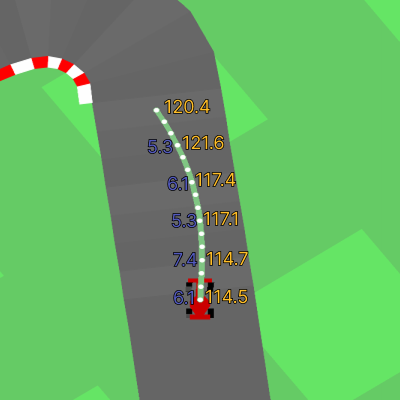}
\end{subfigure}
\hfill
\begin{subfigure}[t]{0.49\linewidth}
    \centering
    \includegraphics[width=\linewidth]{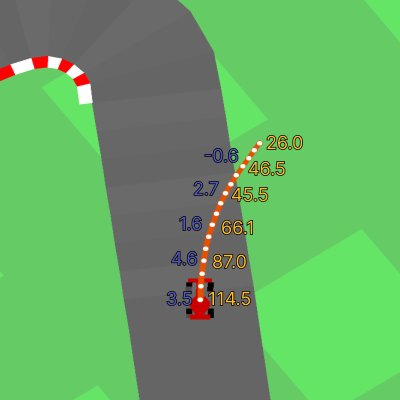}
\end{subfigure}
\vspace{-1.5em}
\caption{\texttt{\textbf{Valdi}} is a diffusion model that predicts action-sequence conditioned \indigo{rewards} and \amber{values}, enabling Model Predictive Control. We show its predictions for a \fern{good} and \flame{bad} action sequence in the CarRacing environment~\citep{Towers2024ARXIV}.}
\label{fig:teaser}
\end{wrapfigure}

In parallel, diffusion models~\citep{Ho2020NeurIPS} have emerged as a strong candidate for world modeling, accurately capturing complex distributions over long horizons~\citep{Agarwal2025ArXiv}. They are typically trained in observation (e.g., pixel) space to sidestep representation collapse~\citep{Alonso2024NEURIPS, Ho2022NIPS}, yet their iterative inference is in tension with the low-latency requirements of MPC.

In this work, we bridge this gap with Value Diffusion World Models (\texttt{Valdi}, illustrated in Figure~\ref{fig:teaser}), an algorithm that trains a latent diffusion dynamics model to predict value functions, end-to-end and in an online control loop, in the style of TD-MPC~\citep{Hansen2022ARXIV, Hansen2023ARXIV}. %
\section{Related Work}
\label{sec:related_work}

\paragraph{Diffusion World Models}
Both pixel-space and latent world models have been used for training Reinforcement Learning (RL) agents inside of them, often referred to as model-based RL~\citep{Ha2018ARXIV, Hafner2020ICLR, Hafner2023ARXIV}. 
In this setting, diffusion models have emerged as a state-of-the-art approach for world modeling~\citep{Alonso2024NEURIPS, Hafner2025ARXIV}.
Originally introduced for generating high dimensional natural data such as images or videos, diffusion models are able to represent highly complex, multi-modal data distributions, making them a natural choice for environments with uncertain state transitions~\citep{Agarwal2025ArXiv, Gao2024NeurIPS, Yang2025NEURIPS}.
Although these methods demonstrate high visual fidelity, autoregressive image generation is hard to apply to planning, since diffusing high-dimensional data is prohibitively expensive. 
Further, the restriction to input space makes it unclear how to adopt multi-sensor setups, e.g., in autonomous vehicles, where a variety of sensors as well as proprioception are used~\citep{Chen2024PAMI}.
In contrast to prior pixel-space diffusion world models  like DIAMOND~\citep{Alonso2024NEURIPS}, \texttt{Valdi} diffuses compact latent states, enabling both compute-efficient control and multiple input signal types.

\paragraph{Temporal Difference Learning for Model Predictive Control (TD-MPC)}
This approach, which is most closely related to ours, pioneered the learning of a \textit{Task-Oriented Latent Dynamics Model} (TOLD) for online planning in a latent space~\citep{Hansen2022ARXIV, Hansen2023ARXIV}.
Given an observation $\bs_t$, its latent representation $\cyan{\bz_t} = E_\theta(\bs_t)$, and the corresponding action taken $\ba_t$, TOLD uses analogous high-level components to our model in Section~\ref{sec:method}: a representation model, a dynamics model, a reward model, and a value model (details in \supp{supp:preliminaries}).
The main architectural difference is the dynamics model: TD-MPC predicts a single next latent state $\cyan{\bz_{t+1}}$ from $(\cyan{\bz_t}, \ba_t)$ and rolls this one-step transition out iteratively to construct imagined trajectories.
At inference time, like TD-MPC, we plan over latent rollouts scored by discounted predicted rewards plus a terminal value, executing only the first action before replanning. We optimize this with the Cross Entropy Method (CEM) rather than TD-MPC's MPPI~\citep{Williams2017}(\supp{supp:planning}).
A limitation of TD-MPC is that its dynamics model is a deterministic MLP that lacks an explicit mechanism for representing ambiguous or multi-modal futures. This is notable given that PlaNet~\citep{Hafner2019ICML}, an early influence on latent planning of this kind, already included a stochastic component.
\texttt{Valdi} builds on the same TD-MPC-style value-guided MPC structure, but replaces the deterministic latent transition model with a diffusion dynamics model, providing this previously missing capability.

\section{Value Diffusion World Models}
\label{sec:method}

\begin{wrapfigure}[20]{r}{0.58\textwidth}
\vspace{-0.4cm}
\centering
\includegraphics[width=\linewidth]{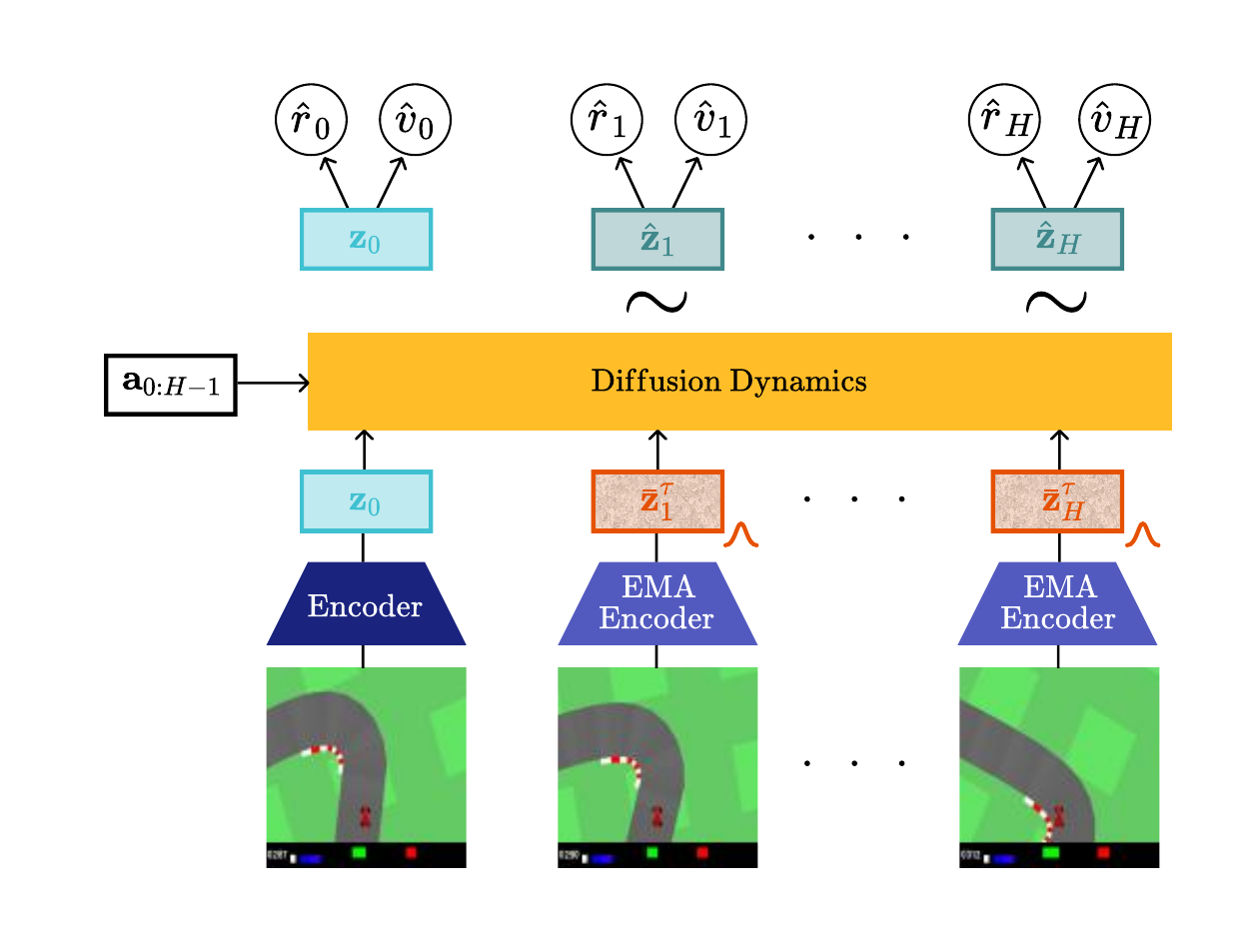}
\caption{\textbf{Method.} \texttt{Valdi} encodes states, conditions diffusion dynamics on an action sequence and noised latents from a target encoder, and uses the denoised latent trajectory for reward and value prediction.}
\label{fig:method}
\end{wrapfigure}

\paragraph{Diffusion-Based TOLD}
Like TD-MPC's TOLD, \texttt{Valdi} consists of four components (Figure~\ref{fig:method}):
\begin{align*}
\textbf{Representation:}\;& \cyan{\bz_t} = E_\theta(\bs_t) \\
\textbf{Dynamics:}\;& \hat{\bu}^{\tau}_{t+1:t+H} = D_\theta(\cyan{\bz_t}, \flame{\bz^{\tau}_{t+1:t+H}}, \ba_{t:t+H-1}, \tau) \\
\textbf{Reward:}\;& \hat{r}_t = R_\theta(\cyan{\bz_t}, \ba_t) \\
\textbf{Value:}\;& \hat{v}_t = V_\theta(\cyan{\bz_t})
\end{align*}

First, rather than rolling out latent states autoregressively, \texttt{Valdi} exploits the diffusion model to generate the entire $H$-step latent trajectory jointly. We use the velocity parameterization of~\cite{Salimans2022ARXIV}, where $\hat{\bu}^{\tau}_{t+1:t+H}$ is the velocity for the noised latent trajectory \flame{$\bz^{\tau}_{t+1:t+H}$} at diffusion timestep $\tau$. Second, we predict state values (with a value head $V_\theta$) rather than action values: \texttt{Valdi}, unlike TD-MPC, has no policy prior, so this simpler formulation suffices (full justification in \supp{supp:preliminaries}). Architecturally, $E_\theta$ is a CNN, $R_\theta$ and $V_\theta$ are MLPs, and $D_\theta$ is a bidirectional encoder-only transformer, totaling $\sim$5M parameters (Supp.~\ref{supp:architecture}).

\paragraph{Training}
We sample trajectories $(\bs_{t:t+H}, \ba_{t:t+H-1}, r_{t:t+H-1})$ from a replay buffer and optimize $\mathcal{L}_{\mathrm{diff}} + \mathcal{L}_{\mathrm{rew}} + \mathcal{L}_{\mathrm{val}}$, with additional regularization, see \supp{supp:pseudocode}. The first component, $\mathcal{L}_{\mathrm{diff}}$, is the standard latent diffusion objective. Crucially, and unlike nearly all latent diffusion models, which freeze their encoders, we \textbf{jointly train} $E_\theta$ with the other components. For the reward and value losses, we obtain one-step denoised predictions from noised state embeddings $\flame{\bar{\bz}^{\tau}_{t+1:t+H}}$ via:
\begin{equation}
    \teal{\hat{\bz}_{t+1:t+H}}
=
\sqrt{\alpha^{\tau}}\, \flame{\bar{\bz}^{\tau}_{t+1:t+H}}
-
\sqrt{1-\alpha^{\tau}}\, \hat{\bu}^{\tau}_{t+1:t+H},
\end{equation}

where \cyan{$\bz$}, \flame{$\bz^\tau$}, and \teal{$\hat{\bz}$} denote clean, noisy, and denoised latents, and a bar marks quantities encoded by an Exponential Moving Average (EMA) target encoder $E_{\bar{\theta}}$. We then apply a TD-MPC-style reward error $\mathcal{L}_{\mathrm{rew}}$ and a temporal difference loss $\mathcal{L}_{\mathrm{val}}$~\citep{Sutton1988ML} to the denoised predictions $\teal{\hat{\bz}_{t+1:t+H}}$ (details and pseudo-code in \supp{supp:pseudocode}). The one-step denoising estimate keeps training tractable and matches our single-step inference for low-latency MPC; we revisit this choice in Section~\ref{sec:experiments}.

\paragraph{Inference}
Our Cross Entropy Method solver~\citep{Rubinstein2004} maximizes discounted returns $\gamma^H V_\theta(\teal{\hat{\bz}_{H}}) + \sum_{h=0}^{H-1} \gamma^h R_\theta(\teal{\hat{\bz}_{h}}, \ba_h)$ over $\ba_{0:H-1}$ and executes the first action $\ba^*_0$ (Supp.~\ref{supp:planning}).

\section{Experiments}
\label{sec:experiments}

As a preliminary proof of concept, all of our experiments use a modified version of the well studied CarRacing environment (see \supp{supp:environment} for details). We are interested in the following questions:
(1) How does the \textbf{performance} of \texttt{Valdi} compare to a standard MLP baseline?
(2) Does the diffusion dynamics model retain its capacity for \textbf{multimodal predictions} when trained jointly with the rest of the system?
(3) How does the choice of dynamics model affect the \textbf{value function}?
We train all models on a single RTX 4080 GPU, with hyperparameters listed in \supp{supp:hyperparameters}.

\begin{wrapfigure}[13]{r}{0.46\textwidth}
\centering
\vspace{-1.0em}
\includegraphics[width=0.46\textwidth]{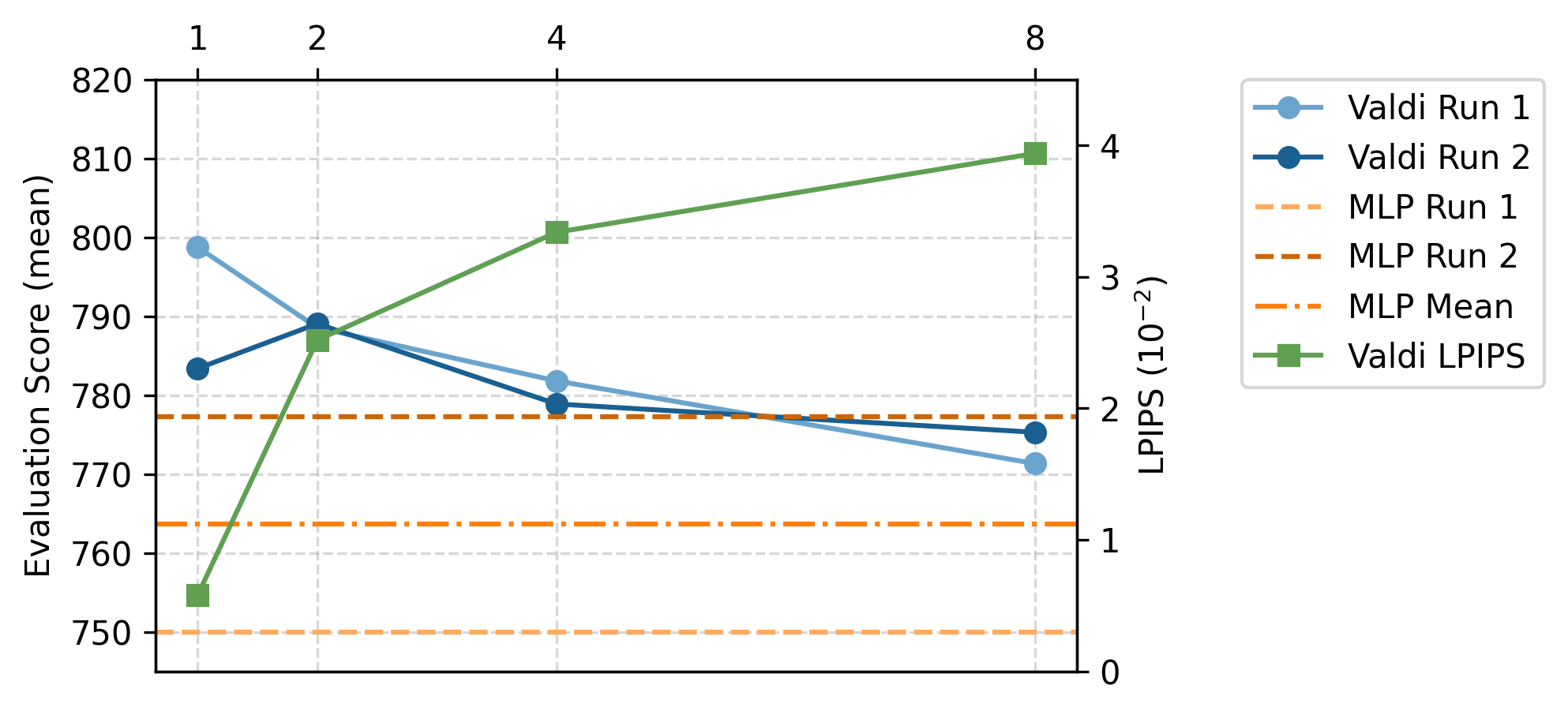}
\vspace{-2.0em}
\caption{\textbf{Performance and multimodality.} Across two runs, more inference diffusion steps do not improve control over our single-step default, but substantially increase the visual variety (LPIPS) among generated futures.}
\label{fig:diff_step_eval}
\end{wrapfigure}

\paragraph{Performance and Multimodality}
We compare \texttt{Valdi} against a baseline that swaps the diffusion dynamics for a deterministic one-step MLP, similar to TD-MPC, with all other parameters identical. Each system is trained twice and evaluated on 100 fixed tracks. Unlike the MLP, \texttt{Valdi} has one additional axis for scaling inference-time compute: the number of diffusion steps. Training and our default evaluation use a single step. In Figure~\ref{fig:diff_step_eval} we explore multi-step inference using a deterministic DDIM sampler~\citep{Song2020ARXIV}.

Further, we extend this experiment to probe predictive variety. We exploit CarRacing's partial observability: the upcoming track is hidden, so given a frame and action sequence, no unique future observation stream is predictable. A deterministic model must therefore output a mean or commit to a single mode, while a diffusion model can in principle represent a distribution over plausible continuations. We train a pixel decoder post-hoc on frozen encoder latents, generate $N=100$ rollouts for several $(\bs_0, \ba_{0:H-1})$ pairs, and report mean pairwise LPIPS~\citep{Zhang2018CVPR} between their decoded final states (visual examples in \supp{supp:multimodal_trajectory_predictions}). A higher LPIPS indicates more variety.

At one diffusion step, \texttt{Valdi} matches the MLP baseline in control within run-to-run variance (additional analysis in \supp{supp:additional_results}). However, its predictive distribution is narrow, typically committing to a single track continuation. Counterintuitively, more diffusion steps slightly degrade control (also within run-to-run variance) while substantially increasing trajectory variety. We attribute the degradation to a training-inference mismatch combined with increased trajectory variance that the CEM solver is ill-equipped to exploit. The result exposes a tension: more diffusion steps yield richer, visually plausible variety, but this variety degrades control performance in this environment. 

\paragraph{Value Function}
Next, we examine how the choice of dynamics model affects the value and reward functions. We evaluate two properties of the value function along imagined rollouts:
(1) \textit{Self-consistency:} does the value function agree with itself across adjacent steps of an imagined rollout, in the sense of a small TD residual?
(2) \textit{Grounding:} does the value of an imagined latent agree with the value of the real latent the agent encounters at the same environment timestep?
Equations for these terms are detailed in the following. 
As in Section~\ref{sec:method}, we use a hat (\teal{$\hat{\bz}$}) for imagined latents and an unadorned symbol (\cyan{$\bz$}) for real, encoder-produced latents. $h$ indexes depth into an imagined rollout and $t$ denotes the environment timestep, with the convention $\teal{\hat{\bz}_0} = \cyan{\bz_t}$ at the start of each rollout.

\paragraph{Self-Consistency} We sample $10{,}000$ real states $\cyan{\bz_t}$ from a post-hoc collected on-policy dataset of each system (i.e., \texttt{Valdi} and the MLP baseline). Starting from $\teal{\hat{\bz}_0} = \cyan{\bz_t}$, we let the planner propose an action sequence $\ba_{0:H-1}$ which the dynamics model uses to imagine a trajectory $\teal{\hat{\bz}_{0:H}}$. We then compute the one-step TD residual $\delta^{\mathrm{TD}}_h$ at each depth. A positive sign for self-consistency indicates that $V(\teal{\hat{\bz}_h})$ underestimates the bootstrap target of $R(\teal{\hat{\bz}_h}, \ba_h) + \gamma V(\teal{\hat{\bz}_{h+1}})$:
\begin{equation}
\delta^{\mathrm{TD}}_h = R(\teal{\hat{\bz}_h}, \ba_h) + \gamma V(\teal{\hat{\bz}_{h+1}}) - V(\teal{\hat{\bz}_h}).
\end{equation}

\paragraph{Grounding} Self-consistency only tells us whether the value function agrees with itself; not whether the imagined latents resemble the real states the agent will actually encounter. To measure this, we roll out our policy in the environment and, at each step, also generate an imagined trajectory from the planner. This gives us, for each environment time $t$ and rollout depth $h$, both an imagined latent $\teal{\hat{\bz}_{h}}$ (starting from $\teal{\hat{\bz}_0} = \cyan{\bz_t}$) and a real latent $\cyan{\bz_{t+h}}$ that the agent reaches $h$ steps later in the real environment. We compute:
\begin{equation}
\delta^{\mathrm{drift}}_h = V(\cyan{\bz_{t+h}}) - V(\teal{\hat{\bz}_h}).
\end{equation}
A negative value indicates an optimistic imagined latent: the world model expects more reward than reality delivers. We aggregate $\delta^{\mathrm{drift}}_h$ over $100$ closed-loop trajectories of $200$ world model steps each.

\begin{wraptable}[11]{r}{0.5\textwidth}
\vspace{-1.35em}
\footnotesize
\centering
\caption{\textbf{Value function diagnostics.} Signed errors ${\delta}$ at each rollout step $h$. \textbf{Bold} marks the model with a lower error, i.e., $|{\delta}|$ closer to zero.}
\label{tab:value_function}
\setlength{\tabcolsep}{3pt}
\begin{tabular}{@{}ll rrrrr@{}}
\toprule
 & & $h{=}0$ & $h{=}1$ & $h{=}2$ & $h{=}3$ & $h{=}4$ \\
\midrule
\multirow{2}{*}{$\delta^{\mathrm{TD}}_{h}$}
 & \texttt{Valdi} & $3.28$           & $2.17$           & $\mathbf{2.00}$  & $\mathbf{1.93}$  & $\mathbf{1.95}$ \\
 & MLP            & $\mathbf{2.05}$  & $\mathbf{2.00}$  & $2.07$           & $2.19$           & $2.23$ \\
\midrule
 & & $h{=}1$ & $h{=}2$ & $h{=}3$ & $h{=}4$ & $h{=}5$ \\
\midrule
\multirow{2}{*}{$\delta^{\mathrm{drift}}_{h}$}
 & \texttt{Valdi} & $-1.53$          & $-2.11$          & $-2.56$          & $-3.04$          & $\mathbf{-3.52}$ \\
 & MLP            & $\mathbf{-0.62}$ & $\mathbf{-1.21}$ & $\mathbf{-1.93}$ & $\mathbf{-2.76}$ & $-3.65$ \\
\bottomrule
\end{tabular}
\vspace{-1.0em}
\end{wraptable}

We report results in Table~\ref{tab:value_function}. At short rollout depths, \texttt{Valdi} is less accurate than the MLP in terms of both self-consistency and grounding. As depth increases, however, \texttt{Valdi}'s error grows more slowly, and near the planner's bootstrap depth H (self-consistency at h = 4, grounding at h = 5) our model is both more self-consistent and better grounded than the baseline. These gaps are small and within run-to-run variance, but the trend is consistent across both diagnostics, suggesting that the predictions that matter most for planning are the ones at which \texttt{Valdi} outperforms the MLP.

\section{Conclusion}
\label{sec:conclusion}

We present Value Diffusion World Models, an algorithm that combines diffusion and latent world models for inference time planning. 
We show that a latent diffusion dynamics model can be trained jointly with a value function inside a TD-MPC-style online loop, matching MLP control performance while producing multi-modal trajectory predictions.

We see two immediate limitations and corresponding directions for future work.
First, the one-step estimate used at both training and inference is unlikely to scale to environments with substantially more complex dynamics, where multi-step denoising would likely be necessary to recover accurate predictions.
To remain computationally feasible at inference, this may then require research into distillation from multi-step teachers to single-step students.
Second, our results suggest that simply changing the number of inference-time diffusion steps degrades planning performance.
Obtaining a model that is flexible at inference time via different diffusion schedules, e.g., enabling test-time scaling, will likely need a corresponding change in the training procedure.

%
%
%
\begin{comment}
\appendix

\section{The first appendix}
\label{sec:appendix1}
This is an example of an appendix. 

\noindent \textbf{Note:} Appendices appear before the references and are viewed as part of the ``main text'' and are subject to the 8--12 page limit, are peer reviewed, and can contain content central to the claims of the paper. 

\section{The second appendix}
\label{sec:appendix2}
This is an example of a second appendix. If there is only a single section in the appendix, you may simply call it ``Appendix'' as follows:

\section*{Appendix}
%
This format should only be used if there is a single appendix (unlike in this document).

\subsubsection*{Acknowledgments}
\label{sec:ack}
Use unnumbered third level headings for the acknowledgments. All acknowledgments, including those to funding agencies, go at the end of the paper. Only add this information once your submission is accepted and deanonymized. The acknowledgments do not count towards the 8--12 page limit.
\end{comment}
%
%
%

%
%
%
\bibliography{main}
\bibliographystyle{rlj}

\beginSupplementaryMaterials

\setcounter{page}{1}
\setcounter{section}{0}
\renewcommand{\thesection}{\Alph{section}}

\section{Preliminaries}
\label{supp:preliminaries}

\subsection{TD-MPC}
TD-MPC trains a \textit{Task-Oriented Latent Dynamics Model} (TOLD), which is then used for inference-time planning in latent space~\citep{Hansen2022ARXIV, Hansen2023ARXIV}. The TOLD model consists of a representation model $E_\theta$ that maps observations to latent states, a dynamics model $D_\theta$ that predicts the next latent state conditioned on the current latent state and action, a reward function $R_\theta$, an action-value function $Q_\theta$, and a prior policy $\pi_\theta$.

The TOLD components are trained jointly with reward prediction, temporal-difference learning for the $Q$-function, and a latent consistency objective that encourages self-consistency over multi-step latent rollouts. The dynamics model in TD-MPC is parameterized as a deterministic MLP that directly predicts the next latent state:
\begin{equation}
\bz_{t+1} = D_\theta(\bz_t, \ba_t).
\end{equation}
This deterministic transition model is efficient, but it does not explicitly represent ambiguous futures introduced by partial observability or stochastic, multi-modal environment transitions.

At inference time, TD-MPC uses the learned model inside a standard MPC loop. It searches for an action sequence by solving
\begin{equation}
\ba^*_{0:H-1}
=
\argmax_{\ba_{0:H-1},\, \bz_{0:H}}
\mathbb{E} \left[
\cJ
\right],
\label{eq:mpc_opt_prob}
\end{equation}
with trajectory score
\begin{equation}
\cJ = \gamma^H Q_\theta(\bz_H, \pi_\theta(\bz_H)) + \sum_{t=0}^{H-1} \gamma^t R_\theta(\bz_t, \ba_t),
\label{eq:traj_score_tdmpc}
\end{equation}
subject to the latent dynamics constraint $\bz_{t+1} = D_\theta(\bz_t, \ba_t)$ for each rollout step. As is common in MPC, the planner executes only the first action $a^*_0$ and replans from the next observed environment state. TD-MPC uses Model-Predictive Path Integral (MPPI) control~\citep{Williams2017} as its MPC solver.

\paragraph{Key Distinctions}
Our method differs from TD-MPC in some key aspects that are likely to result in non-negligible performance implications, the largest of which is our omission of the prior policy $\pi_\theta$. TD-MPC and its successors use this prior policy to bootstrap the planning process with a small percentage of action sequences that are generated by interleaving actions proposed by $\pi_\theta$ with predictions from the dynamics model,
\begin{equation*}
\ba_0\sim\pi_\theta(\bz_0),~~\bz_1=D_\theta(\bz_0, \ba_0), ~~\ba_1\sim\pi_\theta(\bz_1),~~\bz_2=D_\theta(\bz_1, \ba_1),
~~\ba_2\sim\pi_\theta(\bz_2), ~\dots
\end{equation*}
which results in better planning efficiency and possibly asymptotic performance. Because our dynamics model jointly predicts all future states conditioned on the full action sequence, this interleaving is not naturally possible, so we do not bootstrap planning in this way.

The secondary purpose of $\pi_\theta$ is to provide an action for the bootstrap value in Equation~\ref{eq:traj_score_tdmpc} and to make $Q_\theta$ trainable via DDPG~\citep{Lillicrap2016ICLR} or SAC~\citep{Haarnoja2018ICML, Haarnoja2018ARXIV}. We sidestep both uses by employing a state-value function $V_\theta$ in place of an action-value function $Q_\theta$, which significantly reduces code complexity and removes moving parts from training. We do not formally ablate this choice in the present work, as the focus is the dynamics model and its effect on the resulting state values; in early exploration we found no meaningful performance difference between $Q_\theta$ and $V_\theta$ implementations. Crucially, we also use $V_\theta$ for the MLP baseline in Section~\ref{sec:experiments} rather than switching to a $Q_\theta$-plus-$\pi_\theta$ setup, to keep the comparison as clean as possible. Our method differs from TD-MPC in other, smaller respects (e.g., exploration-schedule parameters) that are unlikely to have a significant effect.

\subsection{Diffusion Models}
We train the dynamics model with the velocity-prediction diffusion objective
\begin{equation}
\mathcal{L}_{\mathrm{diff}} =
\left\|
\bu^{\tau}_{t+1:t+H}
-
\hat{\bu}^{\tau}_{t+1:t+H}
\right\|_2, \quad \text{where}
\end{equation}
\begin{align*}
\bu^{\tau}_{t+1:t+H}
&=
\sqrt{\alpha^{\tau}}\, \boldsymbol{\epsilon}
-
\sqrt{1-\alpha^{\tau}}\, \cyan{\bar{\bz}_{t+1:t+H}}
\\[0.5em]
\hat{\bu}^{\tau}_{t+1:t+H}
&=
D_\theta\!\left(
\cyan{\bz_t}, \flame{\bar{\bz}^{\tau}_{t+1:t+H}}, \ba_{t:t+H-1}, \tau
\right)
\\[0.5em]
\flame{\bar{\bz}^{\tau}_{t+1:t+H}}
&=
\sqrt{\alpha^{\tau}}\, \cyan{\bar{\bz}_{t+1:t+H}}
+
\sqrt{1-\alpha^{\tau}}\, \boldsymbol{\epsilon}.
\end{align*}
Here $\boldsymbol{\epsilon} \sim \mathcal{N}(0, I)$, and $\alpha^{\tau} = \prod_{i=1}^{\tau}(1-\beta_i)$ is the cumulative product of the standard linear variance schedule~\citep{Ho2020NeurIPS}, with $\beta$ increasing linearly from $\beta_1 = 1\mathrm{e}{-4}$ to $\beta_T = 2\mathrm{e}{-2}$ over $T = 1000$ steps:
\begin{equation}
\beta_i = \beta_1 + \frac{i-1}{T-1}\left(\beta_T - \beta_1\right), \qquad
\alpha^{\tau} = \prod_{i=1}^{\tau} (1 - \beta_i).
\end{equation}

\section{Architecture Details}
\label{supp:architecture}

Both \texttt{Valdi} and the MLP baseline share the TOLD layout described in Section~\ref{sec:method}: a representation model $E_\theta$, a dynamics model $D_\theta$, a reward model $R_\theta$, and a value model $V_\theta$. The two systems differ \emph{only} in the dynamics model; every other component is identical, isolating the dynamics model as the single variable under study. We summarize all components in Table~\ref{tab:architecture}.

\texttt{Valdi} has $5{,}390{,}230$ trainable parameters and the MLP baseline has $5{,}795{,}363$. We note that the baseline is the slightly \emph{larger} of the two, so any performance parity is not the result of giving \texttt{Valdi} a capacity advantage.

\begin{table}[t!]
\centering
\footnotesize
\caption{Architecture of the shared and dynamics-specific components. The latent dimension is $64$; the per-step action conditioning has dimension $9$ ($3$ raw action dimensions $\times$ an action chunk of $3$, see \supp{supp:planning}). All nonlinearities are $\mathrm{tanh}$-approximated GELU.}
\label{tab:architecture}
\setlength{\tabcolsep}{5pt}
\begin{tabular}{@{}llp{7.6cm}@{}}
\toprule
Component & Type & Details \\
\midrule
$E_\theta$ (visual) & CNN & $5$ conv layers, channels $3\!\to\!16\!\to\!32\!\to\!64\!\to\!128\!\to\!256$, projected to a $256$-d embedding \\
$E_\theta$ (proprio) & MLP & $7\!\to\!256\!\to\!256$ \\
$E_\theta$ (head) & MLP & concat($512$) $\to$ LayerNorm $\to$ $512$ $\to$ $64$ latent \\
\addlinespace
$D_\theta$ (\texttt{Valdi}) & Transformer & $6$ pre-norm blocks, width $256$, FFN $256\!\to\!512\!\to\!256$; final LayerNorm + linear readout to $64$ (velocity) \\
$D_\theta$ (baseline) & Residual MLP & input $73$ $\to$ $384$, $6$ residual blocks (FFN $384\!\to\!768\!\to\!384$), linear readout to $64$ \\
\addlinespace
$R_\theta$ & Residual MLP & input $73$ $\to$ $256$, $2$ residual blocks (FFN $256\!\to\!512\!\to\!256$), linear readout to $1$ \\
$V_\theta$ & MLP ensemble & $2\times$ $\big[64\!\to\!512\!\to\!512\!\to\!1\big]$ (minimum taken over the two heads, see \supp{supp:pseudocode}) \\
\bottomrule
\end{tabular}
\end{table}

\paragraph{Encoder} The representation model encodes the two input modalities separately: the (proprioception-masked) frame through the convolutional stack and the seven-dimensional proprioceptive vector through a small MLP (see \supp{supp:environment} for the modality split). The two $256$-dimensional embeddings are concatenated and passed through a LayerNorm-prefixed projection head to a $64$-dimensional latent.

\paragraph{Dynamics} The \texttt{Valdi} dynamics model is a bidirectional, encoder-only transformer of $6$ pre-norm blocks operating at width $256$, with one token per world-model step. The diffusion timestep $\tau$ is mapped to a $32$-dimensional vector by a learned embedding ($\mathrm{Embedding}(1000, 32)$). Action and diffusion-timestep conditioning are injected by concatenating the per-step action chunk and the timestep embedding to the (noised) latent token along the channel dimension before the first block; a final LayerNorm and linear layer read out the $64$-dimensional velocity per step. The baseline replaces this with a residual MLP of matched depth that maps the current latent and action chunk directly to the next latent.

\paragraph{Reward and Value} The reward model is a residual MLP that takes the latent concatenated with the action chunk ($64 + 9 = 73$). The value model is an ensemble of two identical MLP heads; following common practice we take the elementwise minimum of the two heads when forming TD targets to mitigate value overestimation (\supp{supp:pseudocode}).

\section{Training Details}
\label{supp:pseudocode}

Here we provide a detailed description of our algorithm in the form of pseudocode that closely aligns with our implementation (Algorithm~\ref{alg:diffusion_told_training}).

\paragraph{Training Objective} We optimize the weighted sum of four terms,
\begin{equation}
\mathcal{L}
=
(1-\lambda_{\mathrm{reg}})\big(\lambda_{\mathrm{diff}}\,\mathcal{L}_{\mathrm{diff}}
+ \lambda_{\mathrm{rew}}\,\mathcal{L}_{\mathrm{rew}}
+ \lambda_{\mathrm{val}}\,\mathcal{L}_{\mathrm{val}}\big)
+ \lambda_{\mathrm{reg}}\,\mathcal{L}_{\mathrm{reg}},
\end{equation}
where $\mathcal{L}_{\mathrm{diff}}$ is the velocity diffusion loss (\supp{supp:preliminaries}), $\mathcal{L}_{\mathrm{rew}}$ is a TD-MPC-style reward error, and $\mathcal{L}_{\mathrm{val}}$ is a Temporal Difference loss~\citep{Sutton1988ML}; both $\mathcal{L}_{\mathrm{rew}}$ and $\mathcal{L}_{\mathrm{val}}$ are applied to the one-step denoised latents $\teal{\hat{\bz}_{t+1:t+H}}$. The three task losses are described in Section~\ref{sec:method}; loss weights are listed in \supp{supp:hyperparameters}. We use the minimum over two value heads when forming the TD target, a standard trick to avoid value overestimation.

\paragraph{Latent Collapse and SIGReg} The fourth loss term, $\mathcal{L}_{\mathrm{reg}}$, is a $\mathrm{SIGReg}$ regularizer~\citep{Balestriero2025ARXIV} that encourages an isotropic Gaussian on the latent distribution; we apply it only to the first timestep ($t_0$) of each sampled training trajectory, and refer the reader to~\citep{Balestriero2025ARXIV} for its full definition. We include $\mathrm{SIGReg}$ chiefly so that our reported configuration faithfully reflects our implementation, rather than as a principled component of the method. In our experiments we observed that both \texttt{Valdi} and the MLP baseline learn without collapsing whether or not $\mathrm{SIGReg}$ is enabled; we did not, however, complete full training runs at $\lambda_{\mathrm{reg}} = 0$ due to resource constraints. We therefore make no claim that $\mathrm{SIGReg}$ is either necessary or the operative mechanism preventing collapse, and leave a careful study of which components ground the latent space to future work.

\algrenewcommand\algorithmiccomment[1]{\hfill{\color{indigo}$\triangleright$ \textit{#1}}}

\begin{algorithm}[H]
\caption{Diffusion-Based TOLD Training}
\label{alg:diffusion_told_training}
\begin{algorithmic}[1]
\Require $\theta, \bar\theta$: online and target network parameters
\Require $\eta, \beta, \lambda, \mathcal{B}$: learning rate, EMA coefficient, loss weights, replay buffer
\While{not converged}
    \Statex {\color{indigo}\textit{// Collect episode with MPC policy}}
    \For{$t = 0 \ldots T$}
        \State $\ba_t \sim \Pi^{\mathrm{MPC}}_\theta(\cdot \mid E_\theta(\bs_t))$
            \Comment{Sample action using MPC}
        \State $(\bs_{t+1}, r_t) \sim \mathrm{Env}(\bs_t, \ba_t)$
            \Comment{Step environment}
        \State $\mathcal{B} \leftarrow \mathcal{B} \cup \{\bs_t, \ba_t, r_t, \bs_{t+1}\}$
            \Comment{Store transition in Replay Buffer}
    \EndFor
    \Statex {\color{indigo}\textit{// Update model using data in replay buffer}}
    \For{num updates per episode}
        \State $\bs_{t:t+H},\, \ba_{t:t+H-1},\, r_{t:t+H-1} \sim \mathcal{B}$
            \Comment{Sample trajectory segment from Replay Buffer}
        \State $\bz_{t:t+H},\, \bar{\bz}_{t:t+H}
        = E_\theta(\bs_{t:t+H}),\, E_{\bar\theta}(\bs_{t:t+H})$
            \Comment{Encode observations}
        \State $\tau,\, \boldsymbol{{\epsilon}} \sim U(0,1000),\, \mathcal{N}(0,I)$
            \Comment{Sample diffusion step and noise}
        \State $\bar{\bz}^{\tau}_{t+1:t+H}
        \gets \sqrt{\alpha^\tau}\,\bar{\bz}_{t+1:t+H}
        + \sqrt{1-\alpha^\tau}\,\boldsymbol{\epsilon}$
            \Comment{Forward noising}
        \State $\bu^{\tau}_{t+1:t+H}
        \gets \sqrt{\alpha^\tau}\,\boldsymbol{\epsilon}
        - \sqrt{1-\alpha^\tau}\,\bar{\bz}_{t+1:t+H}$
            \Comment{Velocity target}
        \State $\hat{\bu}^{\tau}_{t+1:t+H}
        \gets D_\theta\!\left(\bz_t, \bar{\bz}^{\tau}_{t+1:t+H}, \ba_{t:t+H-1}, \tau\right)$
            \Comment{Predict denoising direction}
        \State $\hat{\bz}_{t+1:t+H}
        \gets \sqrt{\alpha^\tau}\,\bar{\bz}^{\tau}_{t+1:t+H}
        - \sqrt{1-\alpha^\tau}\,\hat{\bu}^{\tau}_{t+1:t+H}$
            \Comment{Reconstruct latent rollout}
        \State $\mathcal{L}
        \gets (1-\lambda_{\mathrm{reg}})(\lambda_{\mathrm{diff}}\mathcal{L}_{\mathrm{diff}}
        + \lambda_{\mathrm{rew}}\mathcal{L}_{\mathrm{rew}}
        + \lambda_{\mathrm{val}}\mathcal{L}_{\mathrm{val}})
        + \lambda_{\mathrm{reg}}\mathcal{L}_{\mathrm{reg}}$
            \Comment{Total training loss}
        \State $\theta \gets \theta - \eta \nabla_\theta \mathcal{L}$
            \Comment{Update online network}
        \State $\bar\theta \gets (1-\beta)\bar\theta + \beta\theta$
            \Comment{Update target network}
    \EndFor
\EndWhile
\end{algorithmic}
\end{algorithm}

\section{Planning and Inference}
\label{supp:planning}

Due to its simplicity and more widespread use, at inference time, we plan in latent space with a Cross Entropy Method (CEM) solver~\citep{Rubinstein2004} rather than the MPPI solver used by TD-MPC. Starting from the encoded current state $\teal{\hat{\bz}_0} = E_\theta(\bs_t)$, we sample a population of candidate action sequences from a diagonal Gaussian, generate the corresponding $H$-step latent trajectories with the dynamics model, score each by the discounted return
\begin{equation}
\gamma^H V_\theta(\teal{\hat{\bz}_{H}}) + \sum_{h=0}^{H-1} \gamma^h R_\theta(\teal{\hat{\bz}_{h}}, \ba_h),
\end{equation}
refit the sampling distribution to the elite set, and iterate. After the final CEM iteration we execute the first action and replan at the next environment step. The CEM population size, elite count, iteration budget, and initial $(\mu^0, \sigma^0)$ are listed in \supp{supp:hyperparameters}.

\paragraph{Action Chunking} The world model operates at a coarser temporal resolution than the environment: a single world-model step corresponds to $3$ environment steps (action chunk of $3$). Each world-model transition is therefore conditioned on a chunk of $3$ environment actions ($3 \times 3 = 9$ scalar action dimensions, hence the action conditioning of dimension $9$ in \supp{supp:architecture}). With a world-model horizon of $H = 5$, the planner reasons over $5$ world-model steps, equivalent to a $15$-step environment horizon, while only paying for $5$ dynamics evaluations.

\paragraph{Inference Cost} For \texttt{Valdi}, trajectory prediction dominates the planning cycle, so inference cost grows linearly in the number of diffusion steps. At our default of a single diffusion step, the planner runs at over $10$~Hz on a single RTX 4080, and the agent acts at over $30$~Hz in the environment by virtue of action chunking.
The MLP baseline is slightly faster than \texttt{Valdi}; we note, however, that neither approach underwent significant performance optimization.

\section{Environment}
\label{supp:environment}

Here we describe the modifications to the CarRacing environment that we employed in our experiments. We modify the environment so that visual and proprioceptive information are split into two separate inputs. In the original environment the proprioceptive information is rendered as a bar at the bottom of the frame; we black out that section of the visual input and instead expose the proprioceptive information directly as a seven-dimensional state vector. We do this primarily to explore multi-modal inputs for our algorithm, in anticipation of future work. Additionally, we truncate trajectories at $600$ environment frames rather than the usual $1000$.

\section{Hyperparameters}
\label{supp:hyperparameters}

We list all hyperparameters in Table~\ref{tab:hyperparams}. The action-chunking entry is explained in \supp{supp:planning}.

\begin{table}[H]
\centering
\begin{tabular}{ll}
\toprule
Hyperparameter & Value \\
\midrule
Discount factor ($\gamma$) & $0.99$ \\
Seed Trajectories & $10$ \\
Replay buffer size & $1000$ Trajectories \\
Replay sampling technique & Uniform \\
Planning horizon (WM) ($H$) & $5$ \\
Planning horizon (Env) & $15$ \\
Action chunking & $3$ (i.e., model predicts $s_3, s_6,\dots,s_{15}$ conditioned on $a_{0:14}$) \\
Initial parameters ($\mu^0, \sigma^0$) & $(0, 1)$ \\
Population size & $512$ \\
Elite samples & $64$ \\
Iterations & $10$ \\
Inference Diffusion Steps & $1$ \\
Latent dimension & $64$ \\
Learning rate ($\eta$) & $3\mathrm{e}{-4}$ \\
Optimizer ($\theta$) & Adam ($\beta_1 = 0.9, \beta_2 = 0.999$) \\
Reward loss weight ($\lambda_{\mathrm{rew}}$) & $0.01$ \\
Value loss weight ($\lambda_{\mathrm{val}}$) & $0.01$ \\
Diffusion loss weight ($\lambda_{\mathrm{diff}}$) & $1$ \\
Regularization loss weight ($\lambda_{\mathrm{reg}}$) & $0.05$ \\
Exploration schedule ($\epsilon$) & $0.25 \rightarrow 0.05$ \\
 & ($250$ trajectories max $\rightarrow$ $250$ trajectories linear decay) \\
Batch size & $256$ \\
Polyak/EMA coefficient ($\beta$) & $0.005$ \\
Number of updates per trajectory & $60$ (update/data ratio $\tfrac{1}{10}$) \\
$\bar\theta$ update frequency & $1$ \\
\bottomrule
\end{tabular}
\caption{Hyperparameters.}
\label{tab:hyperparams}
\end{table}

\section{Additional Results}
\label{supp:additional_results}

\subsection{Training and Evaluation Returns}
\begin{figure*}[t]
\centering
\begin{subfigure}[t]{0.4\textwidth}
    \vspace{20pt}
    \centering
    \includegraphics[width=\linewidth]{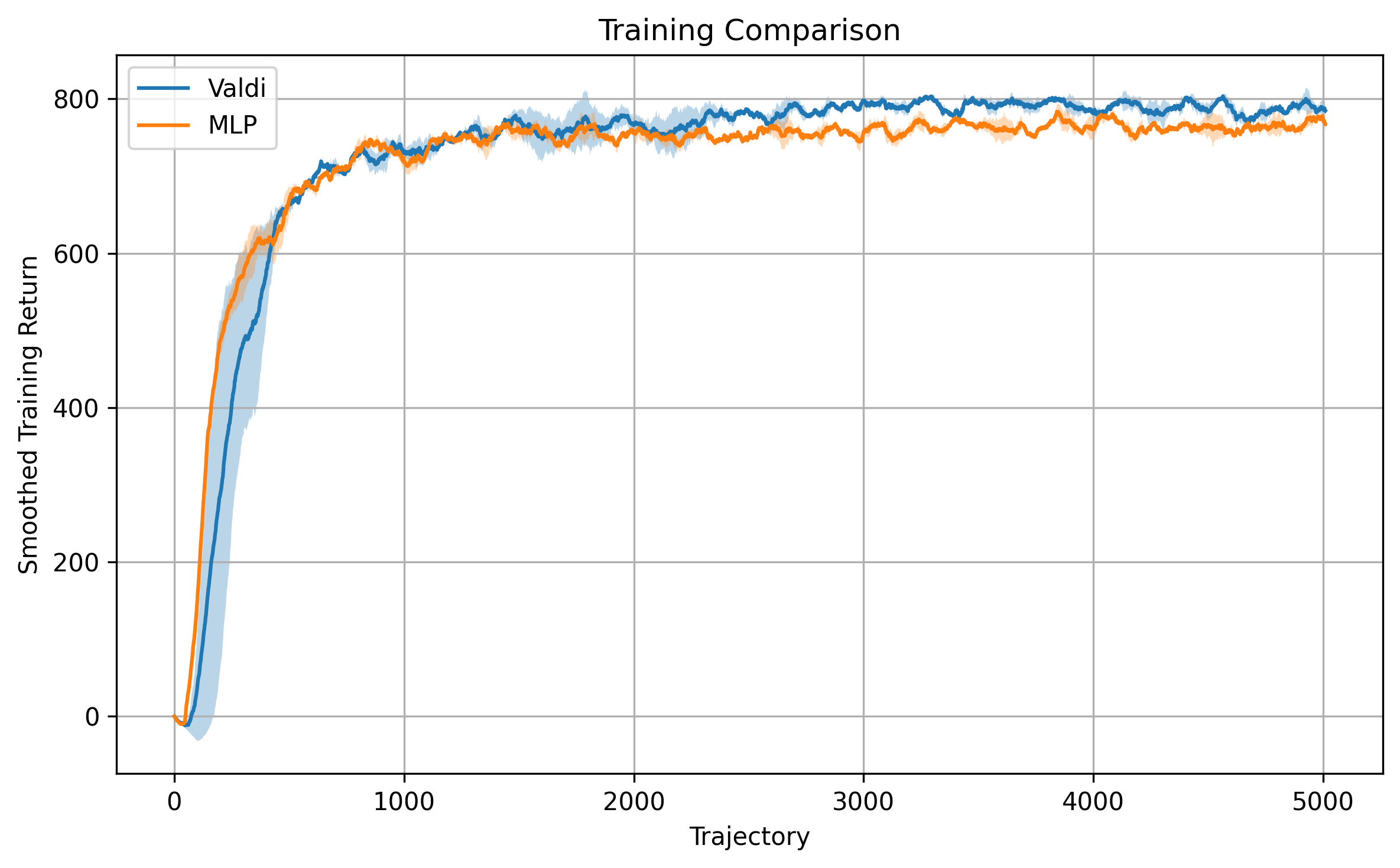}
\end{subfigure}
\hfill
\begin{subfigure}[t]{0.55\textwidth}
    \vspace{0pt}
    \centering
    \includegraphics[width=\linewidth]{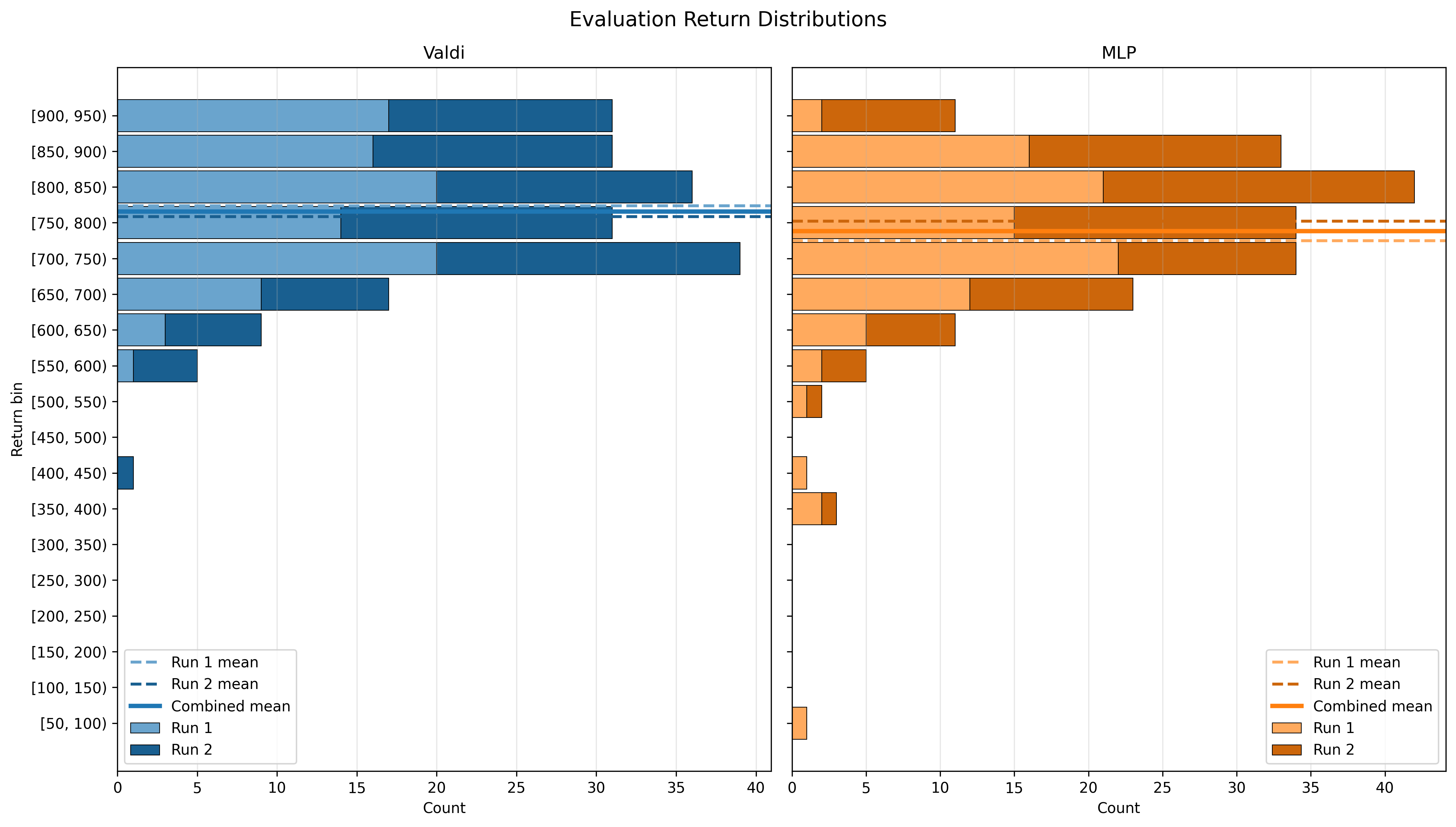}
\end{subfigure}
\caption{The returns that our system and our baseline obtain during training time (left). The performance both runs of both systems obtain at evaluation time (right).}
\label{fig:train_and_eval_results}
\end{figure*}

We report the training-time and evaluation-time returns of both systems in Figure~\ref{fig:train_and_eval_results}. \texttt{Valdi} matches the MLP baseline within run-to-run variance, neither significantly improving nor degrading control performance.

\subsection{Multimodal Trajectory Predictions}
\label{supp:multimodal_trajectory_predictions}

Here we provide an example of our dynamics model's capacity to generate diverse futures from the same start state. We obtain the four trajectories shown by generating $100$ imagined futures from a single starting state, computing pairwise visual similarity between their decoded final states, and selecting the subset of $4$ trajectories that minimizes mutual similarity. We then decode each state of each trajectory independently.

\paragraph{Decoder} The visualizations and the LPIPS measurements in Section~\ref{sec:experiments} rely on a pixel decoder trained \emph{post hoc} on \emph{frozen} encoder latents; the decoder is never used during training or planning. The decoder is a Vision Transformer that reconstructs image patches from a latent. We train one decoder per fully trained encoder, and each decoder is trained on an on-policy dataset collected by rolling out exactly the model whose latents it reconstructs. This ensures each decoder is trained on the latent distribution induced by its own model, so the decoded futures faithfully reflect that model's predictions rather than an out-of-distribution mapping.

\begin{figure}[b]
\centering
\includegraphics[width=\linewidth]{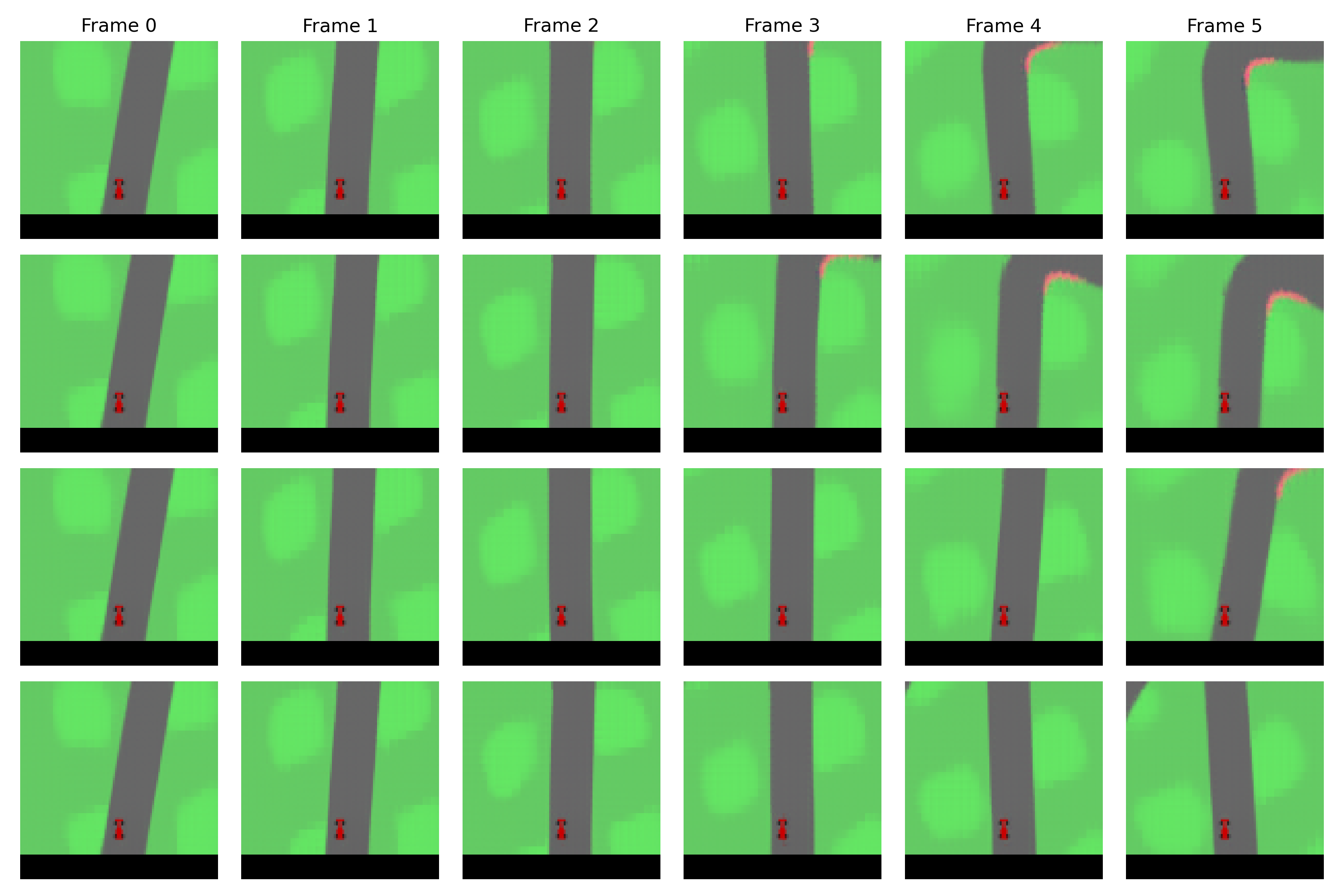}
\caption{Diverse trajectories generated from the same starting state with $8$ diffusion steps. Frame $0$ is identical for all trajectories and serves as the initial state. Frames one through five are imagined by our dynamics model in latent space and then decoded. Our model shows the capacity to generate diverse futures.}
\label{fig:multimodal_trajectory_d8_n_3}
\end{figure}

\end{document}